\documentclass{article}
\usepackage{ijcai24}
\usepackage[numbers]{natbib}

\usepackage{times}
\usepackage{soul}
\usepackage{url}
\usepackage[hidelinks]{hyperref}
\usepackage[utf8]{inputenc}
\usepackage[small]{caption}
\usepackage{amsthm,amsmath,amssymb}
\usepackage{graphicx} 
\usepackage{booktabs}
\usepackage{multirow} 
\usepackage[ruled,vlined]{algorithm2e}
\usepackage{subfigure}
\usepackage[switch]{lineno}
\usepackage{natbib}
\def\ie{\emph{i.e.}}

\title{\vspace{-1.5cm}  Adaptive Principal Components Allocation with the $\ell_{2,g}$-regularized Gaussian Graphical Model  for Efficient Fine-Tuning Large Models}

\author{
Jingjing Zheng$^1$ \and
Yankai Cao$^{2,}$\footnote{Corresponding author}\\
\affiliations
$^1$Department of Mathematics, The University of British Columbia, BC,
Canada\\
$^2$Chemical and Biological Engineering, The University of British Columbia, BC,
Canada\\
\emails
$^1$jjzheng@math.ubc.ca,
$^2$yankai.cao@ubc.ca \\
}

\begin{document}

\maketitle

\begin{abstract}
In this work, we propose a novel Parameter-Efficient Fine-Tuning (PEFT) approach based on Gaussian Graphical Models (GGMs), marking 
 the first application of GGMs to PEFT tasks, to the best of our knowledge. The proposed method utilizes the \(\ell_{2,g}\)-norm to effectively select critical parameters and capture global dependencies. The resulting non-convex optimization problem is efficiently solved using a Block Coordinate Descent (BCD) algorithm. Experimental results on the GLUE benchmark \cite{wang2018glue} for fine-tuning \texttt{RoBERTa-Base} \cite{liu2019roberta} demonstrate the effectiveness of the proposed approach, achieving competitive performance with significantly fewer trainable parameters. The code for this work is available at: \url{https://github.com/jzheng20/Course_projects.git}.

\end{abstract}

\section{Introduction}
Recently, various large models, such as \texttt{BERT}  \cite{devlin2019bert}, \texttt{Roberta}  \cite{liu2019roberta}, \texttt{GPT-3}  \cite{brown2020language}, \texttt{ViT} \cite{dosovitskiy2020image}, and \texttt{PaLM}  \cite{chowdhery2022palm}, have achieved significant success across diverse tasks. Adapting these models for downstream tasks often requires memory-intensive full fine-tuning. To address this, parameter-efficient fine-tuning methods have been developed, such as Prefix-Tuning \cite{li2021prefix,liu2022ptuningv2}, Adapters \cite{houlsby2019parameters}, Sparse Methods \cite{zaken2022bitfit,gao2024parameter}, and Low-rank Adaptation (LoRA) \cite{hu2021lora,meng2024pissa,zhang2023adalora,dettmers2024qlora}.


Among them, LoRA \cite{hu2021lora} has been particularly popular due to its effective reduction of trainable parameters by decomposing the weight increment matrix into two smaller matrices, assuming a low-rank structure. However, it uses a fixed rank  
$r$ across all layers and modules, which limits flexibility. Some modules might require higher ranks to capture detailed information, while others could use lower ranks to save resources. To address this, AdaLoRA \cite{zhang2023adalora} is introduced, in which    an importance scoring mechanism is used to adjusts the rank based on the contribution of singular values to the training, retaining only the most crucial  weights trainable during fine-tuning.




While low-rank methods generally achieve higher parameter efficiency compared to other approaches, they have limitations in capturing global dependencies among parameters. For example, AdaLoRA achieves low-rank adaptation through local adjustments of the increment matrices, without capturing the global dependencies among different parameters. In complex downstream tasks, there may be strong interactions among certain parameters that AdaLoRA fails to effectively model. As a result, some critical interactions might be overlooked, potentially compromising the overall effectiveness of fine-tuning in practical applications.
Graphical models, such as Gaussian Graphical Models \cite{SparseGGM07Friedman,Inter21Cong,Pushing2020Brouard}, are widely used to explore the dependencies and interactions among variables, making them a natural choice for studying parameter relationships in neural networks.


\begin{figure} 
    \centering 
        \includegraphics[width=0.25\textwidth]{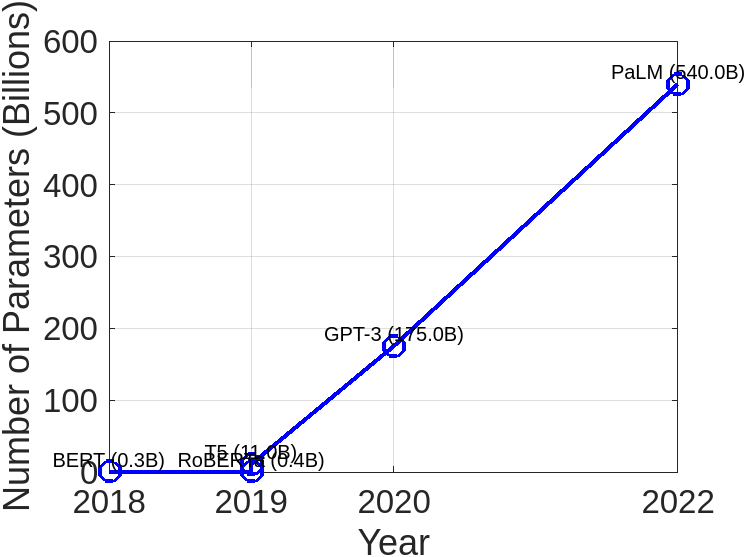} 
     \caption{ Growth of Large Model Parameters (2018–2022).}
     \end{figure} 

In this paper, we aim to introduce a novel Parameter-Efficient Fine-Tuning approach based on Gaussian Graphical Model. We answer the following three questions to  achieve the goal:    
 \begin{itemize} 
         \item \textbf{Question 1:} How to define the nodes of the graph for the PEFT task?
          \item  \textbf{Question 2:} How to define the values of the nodes?
          \item \textbf{Question 3:} How to construct a Gaussian Graphical Model to capture parameter interactions effectively?
 \end{itemize}

Our work has the following two key features:
 \begin{itemize} 
 \item[(1)] Our method utilizes Gaussian Graphical Model  to  capture the interactions among different trainable parameters for specific downstream tasks.  By preserving less critical parameters through freezing and selectively training the more relevant ones, our approach seeks to enhance task-specific adaptation. 
\item[(2)] A non-convex surrogate of $\ell_{2,1}$ norm, $\ell_{2,g}$ norm, is used for the structural sparsity regularization and to enforce sparsity at the node level, helping to select  
the crucial nodes for fine-tuning. To solve the  $\ell_{2,g}$-regularized  Gaussian Graphical Model, we propose an optimization algorithm based on the Block Coordinate Descent (BCD) method. 
 \end{itemize}



  \begin{table}[t]
\centering
\begin{tabular}{|c|l|}
\hline
\textbf{Symbol}        & \textbf{Description}                                \\ \hline
\( A, B, \cdots \)     & Matrices                                   \\ \hline
\( A_j \)              & The \( j \)-th column vector of matrix \( A \)     \\ \hline
\( [A_j]_i \)          & The \( i \)-th element of vector \( A_j \), respectively      \\ \hline
\( A_{i,j} \)          & The \((i, j)\)-th element of matrix \( A \)        \\ \hline
\( a, b, \cdots \)     & Vectors                                    \\ \hline
\(   a_i  \)          & The \( i \)-th element of vector      \( a \), respectively      \\ \hline
$\|a\|_2$& The $\ell_2$ norm of $a$, \ie,  $\|a\|_2=\sum_ia_i^2$\\ \hline
$\|a\|_1$& The $\ell_1$ norm of $a$, \ie,  $\|a\|_1=\sum_i|a_i|$\\ \hline
\end{tabular}
\caption{List of symbols.}
\end{table}\label{Tab_def}

\subsection{Conventions}

In this paper, we use \( W_0 \) to represent a certain weights matrix of a pre-trained  large model and \( \Delta W \) to denote the increment matrix. The symbols and definitions that will be used later in the paper are summarized in the Table \ref{Tab_def}.


\section{The Proposed Methodology}

 \subsection{How to define the nodes and their values?}

In this section, we address the first two questions raised in the Introduction.

For large models, which often consist of billions of parameters, defining nodes on an element-wise basis is impractical. Fortunately, over-parameterized models tend to exhibit low-rank properties in their weight matrices. Figure \ref{lowrank} illustrates the frequency distribution of singular values for the 5-th projection weight matrix of the query in RoBERTa-base and RoBERTa-large. As shown in the figure, most singular values are small and close to zero. Inspired by this observation, we propose selecting the most significant  
$r$ principal components and the bias for each layer as the nodes.

\subsubsection{Node Definition}



 Let  $W_0=USV^{\top}$ represent the weight matrix of a specific layer, where $U$, $S$, and $V$ are obtained  by the  Singular Vale Decomposition (SVD) of $W_0$. 
  Define $A_i=S_{i,i}U_i$ and $B_i=V^{\top}_i$,  where $S_{i,i}$, $U_i$, and $V_i$ correspond to the  
$i$-th singular value, and its associated left and right singular vectors, respectively.  
  We define $r+1$ nodes of that layer as $(A_1, B_1), (A_2, B_2), \cdots, (A_r, B_r), b$, where $b$ is the bias of the layer.  Assuming the model has $h$ layers, we select $r$  principal components per layer results in a total of $h(r+1)$ nodes. 
 
 \subsubsection{Node Value Calculation}
 We use the importance score $s^{(k)}(\cdot)$   \cite{zhang2022platon} to calculate the values of the nodes and  defined as 
 \begin{equation}\label{score}
s^{(k)}(W_{i,j}) = \overline{I}^{(k)}(W_{i,j}) \cdot \overline{U}^{(k)}(W_{i,j}),
 \end{equation}  
 where 
  \begin{equation}
       \overline{I}^{(k)}(W_{i,j}) = \beta_1 \overline{I}^{(k-1)}(W_{i,j}) + (1 - \beta_1) I^{(k)}(W_{i,j}), \notag
  \end{equation} 
   \begin{align}
         \overline{U}^{(k)}(W_{i,j}) = \beta_2 &\overline{U}^{(k-1)}(W_{i,j}) \notag \\
         &+ (1 - \beta_2) \big| I^{(k)}(W_{i,j}) - \overline{I}^{(k)}(W_{i,j}) \big|, \notag
   \end{align} 
  and 
\begin{equation}
    I(W_{ij}) = |W_{ij} \nabla_{W_{ij}} \mathcal{L}|. \notag
\end{equation}
A high $I(W_{i,j}) $ indicates that  $W_{ij}$ has a greater impact on the loss function.

Using the importance score $s^{(k)}(\cdot)$, we calculate the node values as follows:
 \begin{itemize}
     \item For $(A_{i},B_{i})$: 
      \begin{align}\label{v1}
     v^{(k)}\Big((A_{i},B_{i})\Big) = & \frac{1}{2d_1} \sum_{j=1}^{d_1} s^{(k)}([A_{i}]_{j}) \notag \\ 
    &  \qquad   + \frac{1}{2d_2} \sum_{j=1}^{d_2} s^{(k)},([B_{i}]_{j}) 
 \end{align}
where $d_1$ and $d_2$ represent the dimensions of $A_i$ and $B_i$, respectively. 
\item For $b$: 
\begin{equation}\label{v2}
      v^{(k)}(b)=\frac{1}{2d_2} \sum_{j=1}^{d_2} s^{(k)}(b_{j}).
  \end{equation}
 
 \end{itemize}

 Here, $k$ refers to the training step, and different samples are obtained for different $k$.


\begin{figure}[t]
    \centering
    \subfigure[]{ 
        \includegraphics[width=0.22\textwidth]{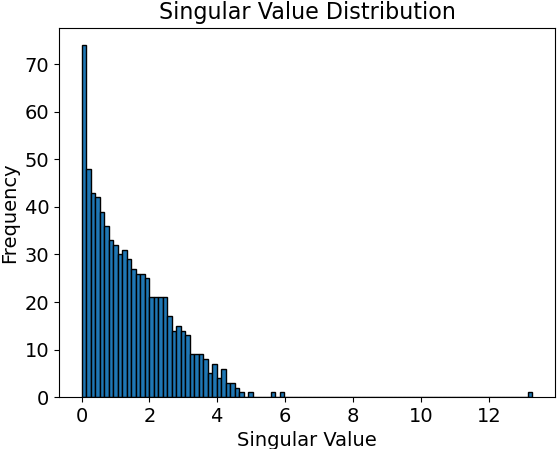}  
    } 
   \subfigure[]{%
        \includegraphics[width=0.22\textwidth]{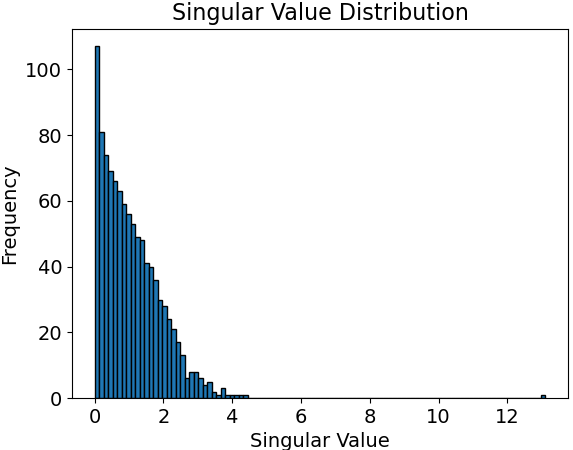} 
    }
    \caption{ Illustration of the low-rank property in the
learned over-parametrized models for (a) Roberta-base (Query) and (b) Roberta-large (Query).}
\end{figure}\label{lowrank}

\subsection{The  Proposed   Gaussian Graphical Model}

\subsubsection{The $\ell_{2,1}$-regularized  Gaussian Graphical Model}

Unlike graphical lasso, which focuses on edge sparsity (\ie, the relationships between individual nodes), the goal of thiswork is to select nodes by learning a structure where a small subset of nodes exhibits stronger interactions with all other nodes. Instead of emphasizing edge connections, we focus on the interaction strength of each node with the remaining nodes. To achieve this, we introduce structural sparsity regularization for the precision matrix \(\Omega \in \mathbb{R}^{n \times n}\).
Specifically, we  measure the interaction strength between node $i$ and all other nodes using \(\|\hat{\Omega}_i\|_2\),
 where  $$\hat{\Omega}_i=[\Omega_{1,i},\Omega_{2,i},\cdots, \Omega_{i-1,i}, \Omega_{i+1,i},\cdots, \Omega_{n,i}]^{\top}\in \mathbb{R}^{n-1}.$$ 
Consequently, we adopt the \(\ell_{2,1}\)-norm of $\hat{\Omega}$: 
$$\|\hat{\Omega}\|_{2,1} = \sum_{i=1}^n \|\hat{\Omega}_i\|_2,$$
instead of the traditional \(\ell_1\)-norm, to better capture these structural interactions. Based on this formulation,  we propose the following Gaussian Graphical Model:
 \begin{equation}\label{l21GGM1}
        \max_{\Omega \succeq 0} \left( \log \det \Omega - \langle \hat{\Sigma}, \Omega \rangle - \tau \|\hat{\Omega}\|_{2,1} \right),
 \end{equation} 
    where $\hat{\Sigma}$ is the sample covariance matrix,  $ \Omega \in \mathbb{R}^{n\times n}$, and $  \|\hat{\Omega}\|_{2,1} = \sum_{j}  \|\hat{\Omega}_j\|_2$. 

\subsubsection{The $\ell_{2,1}$-regularized  Gaussian Graphical Model with ``Important Nodes"}

Although the regularization in \eqref{l21GGM1} effectively promotes structural sparsity, it loses information about the magnitude of the importance scores. This limitation arises because the sample mean is subtracted during the computation of the sample covariance matrix. The sample mean, however, contains crucial information for the Parameter-Efficient Fine-Tuning (PEFT) task. Beyond selecting nodes with strong interactions, it is also essential to prioritize nodes with relatively large values.

Therefore, we introduce the concept of ``important nodes", which are nodes with a significant impact on the loss function. Let \(\mathbb{I} = \{i_1, i_2, \cdots, i_h\}\) represent the set of ``important nodes." We  measure  the interaction strength between node $i$ (\(i \notin \mathbb{I}\)) and the nodes in \(\mathbb{I}\) using \(\|\bar{\Omega}_i\|_2\), where 
$$\bar{\Omega}_i = [\Omega_{i_1, i}, \Omega_{i_2, i}, \cdots, \Omega_{i_h, i}]^{\top} \in \mathbb{R}^h.$$
Based on this concept, we derive a modified   \(\ell_{2,1}\)-regularization term: $$\|\bar{\Omega}\|_{2,1} = \sum_{j \notin \mathbb{I}} \|\bar{\Omega}_j\|_2,$$
and proposed the following improved model: 
 \begin{equation}\label{l21GGM2}
        \max_{\Omega \succeq 0} \left( \log \det \Omega - \langle \hat{\Sigma}, \Omega \rangle - \tau \|\bar{\Omega}\|_{2,1} \right).
 \end{equation} 
 
In thiswork, the set \(\mathbb{I}\) is determined by selecting the nodes with the highest values in the sample mean.

     \begin{table}[t]
\centering 
\begin{tabular}{|c|c|}
\hline
\textbf{Name} & \( g(x)\) \\ \hline
\(\ell_p\) \nocite{frank1993statistical}   & \(x^p, \, 0 < p < 1\) \\ \hline
Geman  \nocite{geman1995nonlinear}  & \(\frac{x}{x + \epsilon}\) \\ \hline
Laplace \nocite{trzasko2008highly} & \((1 - \exp(-\frac{x}{\gamma}))\) \\ \hline
LOG \nocite{malioutov2013iterativelogthresholding}  & \(\log(\gamma + x)\) \\ \hline
Logarithm \nocite{friedman2012fast}  & \(\frac{1}{\log(\gamma + 1)} \log(\gamma x + 1)\) \\ \hline
ETP  \nocite{gao2011feasible} & \(\frac{1 - \exp(-\gamma x)}{1 - \exp(-\gamma)}\) \\ \hline
\end{tabular}
\caption*{\textbf{Table 1:} Examples of surrogate functions of $\ell_0$ norm.}
\end{table}

\subsection{The $\ell_{2,g}$-regularized  Gaussian Graphical Model}

The \(\ell_1\)-norm of  \(v=[\|\bar{\Omega}_{i_1}\|_2, \|\bar{\Omega}_{i_2}\|_2, \cdots, \|\bar{\Omega}_{i_h}\|_2]\),  serves as the convex relaxation of the  $\ell_0$ norm and is widely used to induce sparsity in $v$ for optimization problems like \eqref{l21GGM2}. However, the \(\ell_1\)-norm often fails to produce truly sparse solutions. Achieving effective sparsity generally requires increasing the regularization parameter, which may lead to over-penalization for the solution.

A common  solution in low-rank and sparse learning is to adopt a non-convex strategy, replacing \(|\cdot|\) with surrogate functions \(g(\cdot)\). This approach balances solvability and effectiveness, enabling genuinely sparse  solutions. Incorporating this modification transforms the optimization problem into a non-convex formulation, expressed as follows:
 \begin{equation}\label{l2pGGM}
        \max_{\Omega \succeq 0} \left( \log \det \Omega - \langle \hat{\Sigma}, \Omega \rangle - \tau \|\bar{\Omega}\|_{2,g} \right),
 \end{equation} 
    where  $  \|\bar{\Omega}\|_{2,g} = \sum_{j \notin \mathbb{I}}  g(\|\bar{\Omega}_j\|_2)$,  
 and $g(x)$ is non-negative, increasing, and concave for $x\geq 0$.


Therefore,  we developed a PEFT method utilizing the   \(\ell_{2,g}\)-regularized Gaussian Graphical Model, as outlined in the Algorithm \ref{Ours}.

\begin{algorithm}[t]
\caption{Adaptive Principal Components Allocation with the \(\ell_{2,g}\)-regularized Gaussian Graphical Model}\label{Ours}
  \textbf{Input:} Pre-trained weight matrices, $r$\\  
 \textbf{1.} Reparameterize the pre-trained weight matrices using SVD and extract the principal components as nodes.\\
  \textbf{2.} Set the extracted nodes as trainable and freeze the residual components. \\
 \textbf{3.} Collect samples during the initial stages of the training process for the large model.\\
\textbf{4.}Compute the sample mean and identify \(\mathbb{I}\), the set of nodes corresponding to the highest values in the sample mean.  \\
 \textbf{5.} Solve the proposed \(\ell_{2,g}\)-regularized Gaussian Graphical Model to obtain \(\Omega^*\).\\
 \textbf{6.} Continue training the nodes in \(\mathbb{I}\) and those with large  \(\|\Omega_i^*\|_2\), while freezing the remaining nodes.
 
\end{algorithm}

\section{Optimization by  Block Coordinate Descent (BCD) for the Proposed $\ell_{2,g}$-regularized Model}

    \begin{algorithm}[h]
    \caption{Block Coordinate Descent}
        \SetAlgoLined
        \KwIn{$\hat{\Sigma} \succeq 0$, $\mathbb{I}$, $\lambda > 0$, $\tau > 0$, $g(\cdot)$}
        \KwOut{$\Omega^{T} \succeq 0$}
        Initialize  $\Omega^{(0)} = \text{diag}(\hat{\Sigma})^{-1}$\;
        \For{each iteration $t=1, \ldots, T$}{
            \textbf{1. Update $\Omega^{(t+1)}$ for given $\Delta^{(t)}$:} 
   $  \Omega^{(t+1)} = \arg \max_{\Omega \succeq 0} \Big( \frac{1}{2\lambda}\log \det \Omega - \langle  \frac{A + A^\top}{2}, \Omega \rangle - \frac{1}{2}\|\Omega\|_F^2 \Big)$, $A = \frac{\hat{\Sigma} - 2\lambda \Delta^{(t)}}{2\lambda}$\;
    \BlankLine
     \textbf{2. Update $\Delta^{(t+1)}$ for given $\Omega^{(t+1)}$:}\\
            \For{ $i=1, \ldots, n$}{
           \eIf{$i \in \mathbb{I}$}{ 
                 $\Delta_{i}=\Omega^{(t+1)}_{i}$\;}{ 
  $\alpha^* = \arg\min_\alpha \big( \frac{1}{2} (\|\bar{\Omega}^{(t+1)}_i\|_2 - \alpha)^2 + \frac{\tau}{2\lambda} g(|\alpha|) \big)$\;
               $\Delta^{(t+1)}_{j,i} =\Omega^{(t+1)}_{j,i}$   for $j\notin \mathbb{I}$, $\bar{\Delta}^{(t+1)}_i = \alpha^* \frac{\bar{\Omega}^{(t+1)}_i}{\|\bar{\Omega}^{(t+1)}_i\|_2}$.}        
            }
        }
    \end{algorithm}

Since the optimization problem in \eqref{l2pGGM} is non-convex and challenging to solve directly, we propose an optimization algorithm based on the Block Coordinate Descent (BCD) method \cite{9916142,honorio2012variable}. To simplify \eqref{l2pGGM} for BCD, we reformulate it as follows.

 We first introduce an auxiliary variable $\Delta$ such that  \(\Delta = \Omega\), leading to the equivalent problem:
 $$
    \max_{\Omega \succ 0, \Delta = \Omega} \Big( \log \det \Omega - \langle \hat{\Sigma}, \Omega \rangle - \tau \|\bar{\Delta}\|_{2,g} \Big).$$  
 
  To remove the equality constraint $ \Delta = \Omega$, we add a penalty term $ - \lambda\|\Omega - \Delta\|_F^2 $, encouraging $ \Delta$ to stay close to $\Omega$.  This leads to the following reformulated problem:
  \begin{equation}\label{sim_problem}
       \max_{\Omega \succ 0} \Big( \log \det \Omega - \langle \hat{\Sigma}, \Omega \rangle - \lambda\|\Omega - \Delta\|_F^2 - \tau \|\bar{\Delta}\|_{2,g} \Big),
  \end{equation} 
  where $\lambda >0$. As $\lambda \longrightarrow +\infty$,     $\Delta$ converges to $\Omega$. 

 Using the BCD framework, we alternately optimize $\Omega$ and $ \Delta$  as follows: 
\[ 
\begin{cases} 
  \Omega^{(t+1)}=  \arg\max_{\Omega \succ 0} \Big( \log \det \Omega - \langle \hat{\Sigma},   \Omega \rangle - \lambda\|\Omega - \Delta^{(t)}\|_F^2 \Big); \\ 
 \Delta^{(t+1)}= \arg\max_{\Delta} \Big( - \lambda\|\Omega^{(t+1)} - \Delta\|_F^2 - \tau \|\bar{\Delta}\|_{2,g} \Big).
\end{cases}
\]

It is important to note that, while the second subproblem is non-convex, it has been  well-studied in the field of low-rank sparse representation. By leveraging the Block Coordinate Descent (BCD) approach, we can effectively decompose a more complex problem \eqref{l2pGGM} into two simpler   subproblems. Below, we detail the solutions to these two subproblems.

 \begin{algorithm}[t]
	\caption{Generalized Accelerating Iterative Algorithm (GAI)  \citep{9916142}} \label{related_final gai}
	\KwIn{A real number $y>0$,  a threshold $\lambda > 0$, and  a tolerance $\tau>0$.}
	\KwOut{$ T_{g}(y, \lambda)=x^*_{\mathrm{G}}$\footnote{$ \mathrm{Solve}(y, \lambda)=\mathop{\arg\min}_{x}  \frac{1}{2}(y-x)^{2}+ \lambda g(x)$}.}
	Let
	\begin{equation}
	\left\{
	\begin{array}{lr}
	f_y(x) = \frac{1}{2}(y-x)^{2}+ \lambda g(x)~,& \\
	J_1(x)=y-\lambda g'(x),~&  \\
	J_2(x)=J_1(x)-\frac{(J_1(J_1(x))-J_1(x))(J_1(x)-x)}
	{J_1(J_1(x))-2J_1(x)+x}~.&
	\end{array}
	\right.\notag
	\end{equation}
	$a_0 \leftarrow \max\{ x| J_1'(x)=1~\mbox{or}~ x = 0\}$.\\
	\eIf{$f'_y(a_0)<0$}{
		\small{\texttt{// Find $\hat{x}_{\mathrm{G}}$ by fixed point iteration}}\\
		Initialize $x_{\mathrm{G}}^{(0)} \leftarrow y$,  
		$t \leftarrow 0$\\
		\While{$|J_1(J_1(x^{(t)}_{\mathrm{G}}))-2J_1(x^{(t)}_{\mathrm{G}})+x^{(t)}_{\mathrm{G}}|>\tau$}{
			$x_{\mathrm{G}}^{(t+1)}=J_2(x_{\mathrm{G}}^{(t)})$ \\
			$t \leftarrow t+1$\\
		}
		$\hat{x}_{\mathrm{G}} =J_1(x_{\mathrm{G}}^{(t)})$
	}{
	return $\hat{x}_{\mathrm{G}} = a_0$\
}
If $ f_y(0) > f_y(\hat{x}_{\mathrm{G}})$, return $x^*_{\mathrm{G}}=\hat{x}_{\mathrm{G}}$; otherwise return $x^*_{\mathrm{G}}=0$.\\ 
\end{algorithm}

\subsection{The Updating of the Precision Matrix}

 For a given $\Delta^{(t)}$, we  update $\Omega^{(t+1)}$ by solving 

 \begin{align}\label{sub1}
      &  \max_{\Omega \succ 0}\Big( \frac{1}{2\lambda}\log \det \Omega - \langle \frac{\hat{\Sigma}}{2\lambda}, \Omega \rangle - \frac{1}{2}\|\Omega - \Delta^{(t)}\|_F^2 \Big) \notag \\
      = & \max_{\Omega \succ 0} \Big(\frac{1}{2\lambda} \log \det \Omega - \langle \frac{\hat{\Sigma}}{2\lambda}-\Delta^{(t)}, \Omega \rangle - \frac{1}{2}\|\Omega\|_F^2 \Big) \notag \\
      = &     \max_{\Omega \succ 0} \Big( \frac{1}{2\lambda}\log \det \Omega - \langle A_s, \Omega \rangle - \frac{1}{2}\|\Omega\|_F^2 \Big), 
 \end{align}
    where $A=\frac{\hat{\Sigma}}{2\lambda}-\Delta^{(t)}$ and $A_s=\frac{1}{2}(A+A^{\top})$. 
Since \eqref{sub1} is concave, it can be solved using gradient descent with the following update rule:
  \[
    \Omega^{(k+1)} = \Omega^{(k)} + \eta \big(\frac{1}{2\lambda}( \Omega^{(k)})^{-1} - A_s -   \Omega^{(k)} \big),
    \]
    where \(\eta > 0\) is the learning rate. 
    
   

\subsection{The Updating of the Auxiliary Variable}
For given $\Omega^{(t+1)}$, we   update $\Delta^{(t+1)} $ by solving \begin{align}\label{sub2}
&  \min_{\Delta} \Big(\frac{1}{2}\|\Omega^{(t+1)} - \Delta\|_F^2 + \frac{\tau}{2\lambda}\|\bar{\Delta}\|_{2,g} \Big) \notag \\
=&\min_{\Delta} \Big(\frac{1}{2}\|\Omega^{(t+1)} - \Delta\|_F^2 + \frac{\tau}{2\lambda} \sum_{j \notin \mathbb{I}}g(\|\bar{\Delta}_j\|_2)\Big).
 \end{align}

	

\begin{table*} 
\centering
    \footnotesize
	\begin{tabular}{c|c|ccccccc}
		\hline
        \multirow{2}{*}{Method} &  \# Trainable &   SST-2  & MRPC & CoLA & STS-B  & \multirow{2}{*}{Avg.}\\
        \cline{3-6} &Parameters & Acc. & Acc. & MCC & PCC   &\\
		\hline
		FF &  125M &      94.8 & \textbf{90.2} &\underline{63.6} & \underline{91.2} &  \underline{85.0} \\
        LoRA   &  0.3M  &  \underline{95.1}  & 89.7 & 63.4 & \textbf{91.5} & 84.9 \\
        AdaLoRA   & 0.3M   &  94.5  &  88.7  & 62.0 & 90.5 &  83.9  \\ 
        DyLoRA   & 0.3M  &  94.3  &   89.5  & 61.1 & 91.1 & 84.0 \\
        PiSSA$^{\mathrm{F}}$   &  \underline{0.248M} &  93.6  & 89.3 & 62.2 & 90.0  & 83.8\\   
        Ours-2 &\textbf{0.084M} &   \textbf{95.2}  &   \textbf{90.2} & \textbf{64.4} & 90.4  &  \textbf{85.1} \\ 
		\hline  
	\end{tabular}
    \caption{ The performance comparison of various fine-tuning methods on four datasets (SST-2, MRPC, CoLA, and STS-B) from the GLUE benchmark using the \texttt{RoBERTa-Base} model. Metrics used include accuracy (Acc.) for SST-2 and MRPC, Matthews Correlation Coefficient (MCC) for CoLA, and Pearson Correlation Coefficient (PCC) for STS-B. ``Ours-2'' represents the proposed method with ``important nodes''. Results are averaged over five random seeds $\{0, 11111, 22222, 33333, 44444\}$.}\label{RoBERTa-Base-model}
\end{table*}

 \begin{table} 
\centering
	
	\begin{tabular}{c|cccc}
    	\hline 
     Method  &      MRPC & CoLA  & STS-B   \\
    	\hline  	\hline 
         Ours-1 &    89.2   & 64.8    &    90.2      \\
		\hline  
         Ours-2 &     \textbf{90.2} & \textbf{65.1}   &  \textbf{90.4}    \\ 
         	\hline  
	\end{tabular}
    \caption{Ablation Study:  ``Ours-1'' stands for the proposed method without  ``important nodes''. All results in this table are based on random seed 0.}\label{RoBERTa-Base-model2}
\end{table} 

Since the optimization for each column of \(\Delta\) is independent, we   solve   \(\Delta_i\) separately:
\begin{itemize}
    \item For \(i \in \mathbb{I}\), \[
\min_{\Delta_i} \Big(\frac{1}{2}\|\Omega^{(t+1)}_i - \Delta_i\|_F^2\Big),
\] 
which leads to the solution \(\Delta_{i} = \Omega^{(t+1)}_{i}\). 

       \item For $i \notin \mathbb{I}$, we optimize  \(\Delta_i\) by 
\begin{align}
      \min_{\Delta_i} \Big( \big(\frac{1}{2}\|\bar{\Omega}^{(t+1)}_i - &\bar{\Delta}_i\|_F^2 + \frac{\tau}{2\lambda}g(\|\bar{\Delta}_i\|_{2})\big) \notag \\
      &+\frac{1}{2}\sum_{j \notin \mathbb{I}}(\Omega^{(t+1)}_{j,i}-\Delta_{j,i})^2 \Big)
\end{align}
and it leads to 
\[ 
\begin{cases} 
 \Delta_{j,i}=\Omega^{(t+1)}_{j,i} \text{~for~} j \notin \mathbb{I}; \\ 
 \bar{\Delta}_i=\alpha^* \frac{\bar{\Omega}^{(t+1)}_i}{\|\bar{\Omega}^{(t+1)}_i\|_2}, 
\end{cases}
\]  
     where 
     \begin{align}
         \alpha^* \in T_{g}(\|\bar{\Omega}^{(t+1)}_i\|_2,  \frac{\tau}{2\lambda})=&\arg\min_{\alpha} \Big(\frac{1}{2}(\|\bar{\Omega}^{(t+1)}_i\|_2  \notag \\
         & - \alpha)^2 + \frac{\tau}{2\lambda}g(|\alpha|)\Big). 
     \end{align}

\end{itemize}

  A general  solver for $T_{g}(\|\bar{\Omega}^{(t+1)}_i\|_2,  \frac{\tau}{2\lambda})$, which converges to the global optimal solution with a superlinear rate, is provided in Algorithm \ref{related_final gai}.

\section{Experiments}

In the experiments,   we   compared the proposed methods (Ours-1 and Ours-2) with following SOTA PEFT methods: \textbf{Fully Fine-tuning (FF)},  \textbf{LoRA} \cite{hu2021lora},  \textbf{AdaLoRA} \cite{zhang2023adalora},   \textbf{DyLoRA} \cite{valipour2022dylora}, and \textbf{PiSSA} \cite{meng2024pissa} on Natural Language Understanding tasks, where \textbf{LoRA}, \textbf{AdaLoRA}, \textbf{DyLoRA}, and \textbf{PiSSA} are the matrix-based methods.  The GLUE benchmark  \cite{wang2018glue} 
for the \texttt{RoBERTa-Base} \cite{liu2019roberta} is used for evaluation. Three  key metrics including  Matthew’s
 correlation coefficient (MCC), Pearson correla
tion coefficient (PCC), and accuracy (Acc.)  are used to evaluate the performance of different   fine-tuned models for  CoLA, STS-B, and all other tasks, respectively.  Consistent with \cite{gao2024parameter}, we limit the training to a maximum of 100 epochs and select the best-performing epoch for each run. All experimental results are summarized in Tables \ref{RoBERTa-Base-model}-\ref{RoBERTa-Base-model2}. The best results for each case are highlighted in bold, while the second-best results are underlined. From the results shown in Table \ref{RoBERTa-Base-model}, it is evident that the proposed method achieves comparable performance while using significantly fewer trainable parameters, demonstrating the parameter efficiency of the proposed approach. Furthermore, the comparison between Ours-1 and Ours-2 in Table \ref{RoBERTa-Base-model2} highlights the importance of the introduced concept of ``important nodes."

\section{Conclusion}

In this work, we proposed a novel approach to Parameter-Efficient Fine-Tuning (PEFT) by  using Gaussian Graphical Models. To the best of our knowledge, the first attempt to apply this methodology in the context of PEFT tasks. Our method demonstrates significant parameter efficiency, utilizing far fewer trainable parameters than existing methods while maintaining competitive performance. This highlights the potential of Gaussian Graphical Models in achieving both effectiveness and efficiency in fine-tuning large models.

This work is still ongoing, we plan to conduct further ablation studies and extend our experiments to larger models to strengthen and validate our findings. Additionally, going beyond focus on backpropagation, we aim to explore the contribution of each node during forward propagation  to deepen our understanding and enhance the interpretability of large models.

\bibliographystyle{named}
\bibliography{ijcai24}

\end{document}